\documentclass{article}

\PassOptionsToPackage{numbers}{natbib}

\usepackage[final]{neurips_2020}

\usepackage[utf8]{inputenc} 
\usepackage[T1]{fontenc}    
\usepackage{hyperref}       
\usepackage{amssymb}
\usepackage{url}            
\usepackage{booktabs}       
\usepackage{amsfonts}       
\usepackage{nicefrac}       
\usepackage{microtype}      
\usepackage{xcolor}         
\usepackage{times}
\usepackage{amsmath, amssymb}
\usepackage{mathtools}
\usepackage{latexsym}
\usepackage{todonotes}
\usepackage{comment}
\usepackage{tabto}
\usepackage{soul}
\usepackage{multicol,multirow}
\usepackage{graphicx}
\usepackage{verbatim}
\usepackage{authblk}

\def\BibTeX{{\rm B\kern-.05em{\sc i\kern-.025em b}\kern-.08em
    T\kern-.1667em\lower.7ex\hbox{E}\kern-.125emX}}
    
\usepackage{color, colortbl}    
\usepackage{amssymb}
\usepackage{algorithm}
\usepackage[noend]{algpseudocode}
\usepackage{xcolor}

\usepackage{graphicx}
\usepackage{tabularx}
\usepackage{soul}
\usepackage{multirow}
\usepackage{hhline}

\usepackage{epstopdf}
\usepackage[utf8]{inputenc}

\usepackage{hyperref}
\usepackage{xstring}

\usepackage{xcolor}

\author[1]{\textbf{Yongxin Zhou}}
\author[1]{\textbf{François~Portet}}
\author[1]{\textbf{Fabien Ringeval}}

\affil[1]{Univ. Grenoble Alpes, CNRS, Inria, Grenoble INP, LIG, 38000 Grenoble, France  \qquad}

\title{Effectiveness of French Language Models on Abstractive Dialogue Summarization Task}

\begin{document}

\maketitle

\begin{abstract}
Pre-trained language models have established the state-of-the-art on various natural language processing 
tasks, including dialogue summarization, which allows the reader to quickly access key information from long conversations in meetings, interviews or phone calls. 
However, such dialogues are still difficult to handle with current models because the spontaneity of the language involves expressions that are rarely present in the corpora used for pre-training the language models. 
Moreover, the vast majority of the work accomplished in this field has been focused on English. 
In this work, we present a study on the summarization of spontaneous oral dialogues in French using several 
language specific pre-trained models: BARThez, and BelGPT-2, as well as multilingual pre-trained models: mBART, mBARThez, and mT5. 
Experiments were performed on the DECODA (Call Center) dialogue corpus whose task is to generate abstractive synopses from call center conversations between a caller and one or several agents depending on the situation. Results show that the BARThez models offer the best performance far above the previous state-of-the-art on DECODA. We further discuss the limits of such pre-trained models and the challenges that must be addressed for summarizing spontaneous dialogues.
\end{abstract}

\section{Introduction}

The task of automatic text summarization consists of presenting textual content in a condensed version that retains the essential information. Recently, the document summarization task has seen a sharp increase in performance due to pre-trained contextualized language models \cite{liu-lapata-2019-text}. Unlike documents, conversations involve multiple speakers, are less structured, and are composed of more informal linguistic usage \cite{sacks1978simplest}.

Research on dialogue summarization have considered different domains such as summarization of meetings \cite{Buist04automaticsummarization,wang-cardie-2011-summarizing,oya-etal-2014-template} and phone calls / customer service \cite{tamura-etal-2011-extractive,Stepanov2015,LiuKDD19}. 

The introduction of the Transformer architecture \cite{attention_2017} has encouraged the development of large-scale pre-trained language models. Compared to RNN-based models, transformer-based models have deeper structure and more parameters, they use self-attention mechanism to process each token in parallel and represent the relation between words. Such model have pushed the performances forward in automatic summarization. For instance, on the popular benchmark corpus CNN/Daily Mail \cite{hermann2015teaching}, \cite{liu-lapata-2019-text} explored fine-tuning BERT \cite{devlin-etal-2019-bert} to achieve state-of-the-art performance for extractive news summarization, and BART \cite{lewis-etal-2020-bart} has also improved generation quality on abstractive summarization. However, for dialogue summarization, the impact of pre-trained models is still not sufficiently documented. In particular, it is unclear if the pre-training dataset mismatch will negatively impact the summarization capability of these models. 

In terms of resources, the dialogue summarization task, has been using well-known corpora such as CNN-DailyMail \cite{hermann2015teaching}, Xsum \cite{xsum-emnlp},  SAMSum chat-dialogues Corpus \cite{gliwa-etal-2019-samsum} and AMI Meeting Corpus \cite{AMI_meeting_corpus}. However, most of the available data sets are English language corpora.

In this paper, we present a study to evaluate the impact of pre-trained language models on a dialogue summarization task for the French language. We had evaluated several pre-trained models of French (BARThez, BelGPT-2) and multilingual pre-trained models (mBART, mBARThez, mT5). The experiments were performed on the DECODA (Call Center) dialogue corpus whose task is to generate abstractive synopses from call centre conversations between a caller and one or more agents depending on the situation. Our experimental results establish a new state-of-the-art with the BARThez models on DECODA. The results show that all BART-based pre-trained models have advanced the performance, while the performance of mT5 and BelGPT-2 is not satisfactory enough.

\section{Related Work}

\subsection{Dialogue Summarization}
As for summarization tasks, there are two different methods: extractive methods and abstractive methods. Extractive methods consist of copying directly content from the original text, usually sentences. Abstractive methods are not restricted to simply selecting and rearranging content from the original text, they are able to generate new words and sentences. 

With the development of dialogue systems and natural language generation techniques, the resurgence of dialogue summarization has attracted significant research attention, which aims to condense the original dialogue into a shorter version covering salient information \cite{survey_dialogue_summarization_2021}. Contrary to well-formed documents, transcripts of multi-person meetings / conversations have various dialogue turns and possible extended topics, summaries generated by NLG models may thus be unfocused on discussion topics. 

\cite{survey_dialogue_summarization_2021} provide an overview of publicly available research datasets and summarize existing works as well as organize leaderboards under unified metrics, for example, leaderboards of meeting summarization task on AMI \cite{AMI_meeting_corpus} and ICSI \cite{ICSI_meeting_corpus}, as well as a leaderboard of chat summarization task on SAMSum \cite{gliwa-etal-2019-samsum}. Summarized by them, 12 out of 15 major datasets for dialogue summarization are in English. 

Taking the chat summarization task on SAMSum dataset as an example, the results showed that the Pre-trained Language Model-based Methods are skilled at transforming the original chat into a simple summary realization \cite{survey_dialogue_summarization_2021}: the best ROUGE-2 score of the Pre-trained Language Model-based Methods is 28.79 \cite{feng-etal-2021-language}, compared to the best ROUGE-2 score of Abstractive Methods at 19.15 \cite{zhao-etal-2020-improving} and the best ROUGE-2 score of Extractive Methods at 10.27 (LONGEST-3), deep neural models and especially pre-trained language models have significantly improved performance.

According to the domain of input dialogue, summarization on DECODA call-center conversations is of customer service summarization type, and it is task-oriented. Previous work has been done on the DECODA dialogue summarization task: \cite{trione-2014-extraction,Trione_2017} used extractive methods and \cite{Stepanov2015} used abstractive approaches. \cite{Stepanov2015} used domain knowledge to fill hand-written templates from entities detected in the transcript of the conversation, using topic-dependent rules. Although pre-trained models have shown superior performance on English, however we have few information about how such models would behave for other languages.

\subsection{French Language Models and Multilingual Models}

If the majority of language models were pre-trained on English-language texts, there have recently been the release of many French oriented pre-trained language models. Furthermore, multilingual pre-trained models have also emerged and showed very good generalisation capability in particular with under-resource languages. A brief summary of the French language models and multilingual models that we used can be found in Table~\ref{table:French Language Models and Multilingual Models}. 

Regarding French pre-trained models, shortly following the release of the BERT-like models: CamemBERT \cite{martin-etal-2020-camembert} and FlauBERT \cite{le-etal-2020-flaubert-unsupervised}, BARThez \cite{kamal-eddine-etal-2021-barthez} was proposed. It is the French equivalent of the BART model \cite{lewis-etal-2020-bart} which uses a full transformer architecture (encoder-decoder). Being based on BART, BARThez is particularly well-suited for generative tasks. 

For generation, there are also French GPT-like models. For instance, BelGPT-2 \cite{louis2020belgpt2}, which is a GPT-2 model pre-trained on a very large and heterogeneous French corpus ($\sim$60Gb). Other French GPT-like models have been proposed in the literature such as PAGnol \cite{PAGnol2021} which is a collection of large French language models, GPT-fr \cite{simoulin2021modele} which uses the OpenAI GPT and GPT-2 architectures \cite{Radford2018ImprovingLU} or even the commercial model Cedille \cite{muller2022cedille}. We used BelGPT-2 in our experiment since it is open-source and a good trade-off between lightweight model and training size. 

Another way to address non-English language tasks is to benefit from multilingual models that were pre-trained on datasets covering a wide range of languages. For example, mBERT \cite{devlin-etal-2019-bert}, mBART \cite{liu-etal-2020-multilingual-denoising}, XLM-R \cite{conneau-etal-2020-unsupervised_XLM-R}, and mT5 \cite{xue-etal-2021-mt5} are popular models of this type, which are respectively the multilingual variants of BERT \cite{devlin-etal-2019-bert}, BART \cite{lewis-etal-2020-bart}, RoBERTa \cite{liu2019roberta} and T5 \cite{T5_RaffelSRLNMZLL20}.

In this study, we will consider mBART, mT5, as well as mBARThez. The latter was designed by the same authors as BARThez \cite{kamal-eddine-etal-2021-barthez} by fine-tuning a multilingual BART on the BARThez training objective, which boosted its performance on both discriminative and generative tasks. mBARThez is thus the most French-oriented multilingual model, this is why we have included it in our experiments.

\begin{table*}[!h]
\begin{center}
\begin{tabularx}{\textwidth}{|X|X|X|X|}

      \hline
      Model&Description& Pre-training corpus & \#params\\
      \hline
      BARThez\cite{kamal-eddine-etal-2021-barthez} & sequence to sequence pre-trained model & French part of CommonCrawl, NewsCrawl, Wikipedia and other smaller corpora (GIGA, ORTOLANG, MultiUn, EU Bookshop) & 165M (architecture: BASE, layers: 12)\\
      \hline
      BelGPT2\cite{louis2020belgpt2} & a GPT-2 model pre-trained on a very large and heterogeneous French corpus ($\sim$60Gb) & CommonCrawl, NewsCrawl, Wikipedia, Wikisource, Project Gutenberg, EuroParl, NewsCommentary & Small (124M)\\
      \hline
      mbart.CC25\cite{mBART} & mBART model with 12 encoder and decoder layers & trained on 25 languages' monolingual corpus, Common Crawl (CC25) & 610M\\
      \hline
      mBARThez\cite{kamal-eddine-etal-2021-barthez} & continue the pretraining of a multilingual BART on BARThez' corpus & the same as BARThez' corpus & 458M (architecture: LARGE, layers: 24)\\
      \hline
   mT5\cite{xue-etal-2021-mt5} & a multilingual variant of T5, Encoder-decoder & new Common Crawl-based dataset covering 101 languages, Common Crawl (mC4) & 300M – 13B\\
      \hline

\end{tabularx}
\caption{Summary of French Language Models and Multilingual Models.}
\label{table:French Language Models and Multilingual Models}
 \end{center}
\end{table*}

\section{Summarizing Call Center Dialogues in French}

\subsection{Motivation}
As pre-trained models showed advanced performance on summarization tasks in English, we would like to explore the behavior of French pre-trained models on dialogue summarization task. In order to deal with it, we choose the corpus DECODA \cite{bechet-etal-2012-decoda} and the task of Call Centre Conversation Summarization, which had been set during Multiling 2015 \cite{favre-etal-2015-call}. The objective is to generate abstractive synopses from call center conversations between a caller and one or more agents.

Previous works on DECODA dialogue summarization reported results of extractive methods and abstractive results, our hypothesis is that by using pre-trained language models, we could get better results in this task.

Besides, in call center conversations scenarios, the caller (or user) might express their impatience, complaint and other emotions, which are especially interesting and make DECODA corpus a seldom corpus for further research related to emotional aspects of dialogue summarization.

\subsection{DECODA Corpus and Data Partitioning}
DECODA call center conversations are a kind of task-oriented dialogue. Conversations happened between the caller (the user) and one agent, sometimes there were more than one agent. The synopses should therefore contain information of users' need and if their problems have been answered or solved by the agent.

Top 10 most frequent topics on the DECODA corpus are: Traffic info (22.5\%), Directions (17.2\%), Lost and found (15.9\%), Package subscriptions (11.4\%), Timetable (4.6\%), Tickets (4.5\%), Specialized calls (4.5\%), No particular subject (3.6\%), New registration (3.4\%) and Fare information (3.0\%) \cite{trione-2014-extraction}. Besides, a call lasts between 55 seconds for the shortest to 16 minutes for the longest.

Table~\ref{table:excerpt_decoda_corpus} shows an excerpt of a dialogue. Prefix A (resp B) indicate the employee (resp customer)'s turn. 

\begin{table*}[!h]
\begin{center}
\begin{tabularx}{\textwidth}{|X|X|}

      \hline
      \multicolumn{2}{|c|}{FR\_20091112\_RATP\_SCD\_0826}\\
      \hline
      French & Translated  \\
      \hline
      A: bonjour B: oui bonjour madame B: je vous appelle pour avoir des horaires de train en+fait c' est pas pour le métro je sais pas si vous pouvez me les donner ou pas A: trains B: oui trains oui A: vous prenez quelle ligne monsieur  \dots & \textit{A: hello B: yes hello Mrs B: I'm calling you to get train timetables in fact it's not for the metro I don't know if you can give them to me or not A: trains B: yes trains yes A: which line do you take sir \dots} \\
      \hline

\end{tabularx}
\caption{Excerpt \textit{FR\_20091112\_RATP\_SCD\_0826} of DECODA conversations.}
\label{table:excerpt_decoda_corpus}
 \end{center}
\end{table*}

We used the same data as in the Multiling 2015 CCCS task (Call Centre Conversation Summarization) \cite{favre-etal-2015-call}: they provided 1000 dialogues without synopses (auto\_no\_synopses) and 100 dialogues with corresponding synopses for training. As for test data, there are in total 100 dialogues. We divided randomly the original 100 training data with synopses into training and validation sets, with separately 90 and 10 samples. Note that for each dialogue, the corresponding synopsis file each may contain up to 5 synopses written by different annotators. Data statistics are presented in Table~\ref{table: decoda_statistics}.

\begin{table}[!h]
\begin{center}
\begin{tabularx}{\columnwidth}{|l|X|X|X|}

      \hline
       & \# dialogues & \# synopsis & \# dialogues w/o\\
       &               &              & synopsis\\
      \hline
      train & 90 & 208 & 1000 \\
      \hline
      dev & 10 & 23 & - \\
      \hline
      test & 100 & 212 & - \\
      \hline

\end{tabularx}
\caption{Number of dialogues and synopses for the train/dev/test and unannotated sets of the DECODA corpus.}
\label{table: decoda_statistics}
 \end{center}
\end{table}

\subsection{Processing Pipeline}

\begin{figure*}[!h]
\begin{center}

\includegraphics[width=.9\linewidth]{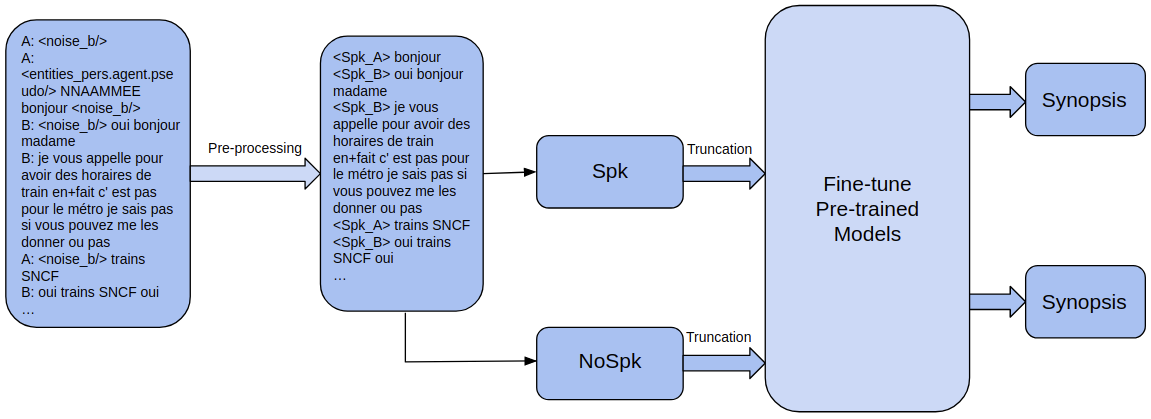} 

\caption{Processing pipeline for dialogue summarization.}
\label{fig:pipline_barthez}
\end{center}
\end{figure*}

The overall process for dialogue summarization using pre-trained language models is depicted in Figure~\ref{fig:pipline_barthez}. Pre-trained languages models have limits to the maximum input length. For instance, BART and GPT-2 based models only accept a maximum input length of 1024 tokens, which pose problems when input sequences are longer than their limits. How to process long dialogue summarization is an ongoing research topic \cite{zhang-etal-2021-exploratory-study}. In this work, since the objective is to compare pre-trained language models, we use the simple technique of truncation in order to meet the length requirement of pre-trained models. Although truncation is not the most advanced approach, in summarization, using the first part of the document is known to be a strong method and hard to beat, in particular in news where the main information is conveyed by the first line. In the case of call center conversations, we can also assume that the most important information is present at the beginning of the conversation since the client initiates the call by stating his / her problems.

\subsubsection{Pre-processing}

Since DECODA was acquired in a real and noisy environment, the DECODA transcripts contain various labels related to the speech context, such as $<$noise\_b/$>$, $<$noise\_rire/$>$, $<$noise\_i/$>$ \textit{speech} $<$noise\_i/$>$, etc. There are also  $<$overlap$>$ \textit{speech} $<$/overlap$>$, which indicates the speech overlap of two speakers. Since the models are not processing speech but only transcript, we filtered out all these labels.

Each speaker of the dialogue transcript is preceded by a prefix (e.g. ``A:'' ). Since the target synopsis is always written in a way that makes the problem explicit (e.g. what the user wants or what the user's problem is) and how it has been solved (by the agent), we wanted to check whether the model is able to summarize without knowing the prefix. Hence, two versions of the input were made: \textit{Spk} presents conversations with speakers’ prefix and \textit{NoSpk} presents conversations without the prefix. For this we used special tokens $<$Spk\_A$>$, $<$Spk\_B$>$, \dots to represent a change in the speakers' turns.

\subsubsection{Truncation}

For the BART and T5 architectures, the DECODA data files are prepared in JSONLINES format, which means that each line corresponds to a sample. For each line, the first value is \textit{id}, the second value is \textit{synopsis} and will be used as the summary record, the third value is \textit{dialogue} and will be used as the dialogue record.

Since BART \cite{lewis-etal-2020-bart} is composed of bi-directional BERT encoders and autoregressive GPT-2 decoders, the BART-based models take the dialogues as input from the encoder, while the decoder generates the corresponding output autoregressively. For BART-based models (BARThez, mBART and mBARThez), the maximum total length of the input sequence after tokenization is 1024 tokens, longer sequences will be truncated, shorter sequences will be padded.

As with mT5, the maximum total length of the input sequence used is also 1024 tokens. We used \textit{source\_prefix: ``summarize: "} as a prompt, so that the model can generate summarized synopses.

When fine-tuning BelGPT-2 for the summarization task, our input sequences are \textit{dialogue} with a \textit{$<$sep\_token$>$} and the \textit{synopsis} at the end. For long dialogues, we have truncated the dialogue and ensured that the truncated dialogue with the separate token and synopsis will not exceed the limit of 1024 tokens. Which means, BelGPT-2 has more constraints than BARThez in the length of dialogues for summarization tasks.

\subsubsection{Summarization}

As mentioned before, we split original 100 training data into training and validation sets, we have 90 dialogue examples with annotated synopses for training. For each dialogue, it may have up to 5 synopses written by different annotators, we paired up dialogue with synopses to make full use of available data for training. In this way, we have 208 training examples, 23 validation examples and 212 test examples.

In experiments, we fine-tuned separately pre-trained models and then used models fine-tuned on DECODA summarization task to generate synopses for test examples. When input sequences exceed the maximum length of the model, we performed truncation as well.

\section{Experiments}

\subsection{Experimental Setup}

For all pre-trained models, we used their available models in Huggingface. Experiments for each model were run separately on two versions of the data (\textit{Spk} and \textit{NoSpk}), running on 1 NVIDIA Quadro RTX 8000 48Go GPU server. 

For BARThez, mBART, mBARThez and mT5,  we used the default parameters for summarization tasks, which are: initial learning rate of ${5e-5}$, with a train\_batch\_size and eval\_batch\_size of 4, seed of 42. We used Adam optimizer with a linear learning scheduler.

In details, we fine-tuned BARThez\footnote{\scriptsize \url{https://huggingface.co/moussaKam/barthez}} (base architecture, 6 encoder and 6 decoder layers) for 10 epochs and saved the best one having the lowest loss on the dev set. Each training took approximately 25 minutes. 

mBART\footnote{\scriptsize \url{https://huggingface.co/facebook/mbart-large-cc25}} (large architecture, 12 encoder and decoder layers) was also fine-tuned for 10 epochs, each training took approximately 46 minutes.

Regarding mBARThez\footnote{\scriptsize \url{https://huggingface.co/moussaKam/mbarthez}}, it is also a large architecture with 12 layers in both encoder and decoder. We fine-tuned it for 10 epochs, each training took approximately 38 minutes. 

As for mT5\footnote{\scriptsize \url{https://huggingface.co/google/mt5-small}} (mT5-Small, 300M parameters), we used \textit{source\_prefix: ``summarize: "} as a prompt to make the model generate synopses. mT5 was fine-tuned for 40 epochs, each training took approximately 65 minutes.

Besides, BelGPT-2 \cite{louis2020belgpt2} was fine-tuned for 14 epochs, with a learning rate of ${5e-5}$, gradient\_accumulation\_steps of 32. To introduce summarization behavior, we add \textit{$<$sep\_token$>$} after the dialogue and set the maximum generated synopsis length at 80 tokens. Each training took approximately 4 hours. During synopsis generation, we tried nucleus sampling and beam search, we reported results with top\_k = 10, top\_p = 0.5 and temperature = 1, which were slightly better.

\subsection{Evaluation}

During CCCS 2015 \cite{favre-etal-2015-call}, different systems were evaluated using the CCCS evaluation kit, with ROUGE-1.5.5 and python2.

In our work, models were evaluated with the standard metric ROUGE score (with stemming) \cite{lin-2004-rouge}. We reported ROUGE-1, ROUGE-2 and ROUGE-L\footnote{The ROUGE metric used in Hugging Face is a wrapper around Google Research reimplementation of ROUGE: \url{https://github.com/google-research/google-research/tree/master/rouge}, we used it to report all our results.}, which calculate respectively the word-overlap, bigram-overlap and longest common sequence between generated summaries and references.

\subsection{Quantitative Results}\label{sec:results}
Results on the different pre-trained language models are shown in Table~\ref{table:ROUGE_scores}.

\begin{table*}[!h]
\begin{center}
\begin{tabularx}{\textwidth}{|X|X|l|l|l|}
\hline
\textbf{\#} & \textbf{Models} & \textbf{ROUGE-1} & \textbf{ROUGE-2} & \textbf{ROUGE-L} \\
\hline
\hline
\cite{favre-etal-2015-call} & Baseline-MMR  & - & 4.5 & - \\
\hline
\cite{favre-etal-2015-call}& Baseline-L  & - & 4.0 & -\\
\hline
\cite{favre-etal-2015-call}& Baseline-LB  & - & 4.6 & - \\
\hline
\hline
\cite{favre-etal-2015-call}& NTNU:1  & - & 3.5 & - \\
\hline
\cite{linharespontes:hal-01779304} & LIA-RAG:1  & - & 3.7 & - \\
\hline
\hline
\cite{kamal-eddine-etal-2021-barthez} & BARThez \#Spk & \textbf{35.45} & \textbf{16.85} & \textbf{29.45} \\
\hline
\cite{kamal-eddine-etal-2021-barthez} & BARThez \#NoSpk&  33.01 &  15.30 &  27.92 \\
\hline
\cite{louis2020belgpt2} & BelGPT-2 \#Spk& 17.73 & 5.15 & 13.82 \\
\hline
\cite{louis2020belgpt2} & BelGPT-2 \#NoSpk& 18.06 & 4.55 & 13.15 \\
\hline
\hline
\cite{liu-etal-2020-multilingual-denoising} & mBART \#Spk& 33.76 & 14.13 & 27.66 \\
\hline
\cite{liu-etal-2020-multilingual-denoising}  & mBART \#NoSpk& 31.60 & 13.69 & 26.20 \\
\hline
\cite{kamal-eddine-etal-2021-barthez} & mBARThez \#Spk& 34.95 & 15.86 & 28.85 \\
\hline
\cite{kamal-eddine-etal-2021-barthez}  & mBARThez \#NoSpk& 35.31 & 15.52 & 28.61 \\
\hline
\cite{xue-etal-2021-mt5} & mT5 \#Spk& 23.90 & 8.69 & 20.06 \\
\hline
\cite{xue-etal-2021-mt5}  & mT5 \#NoSpk& 27.82 & 11.08 & 23.36 \\
\hline
\end{tabularx}
\caption{ROUGE-2 performances of the baseline and submitted systems for the multiling challenge 2015 on French DECODA 
and results of ROUGE-1, ROUGE-2 and ROUGE-L scores of the pre-trained language models. The best results are boldfaced.} 
\label{table:ROUGE_scores}
\end{center}
\end{table*}

The table reports the ROUGE-2 scores (Recall only) of the baseline and submitted systems on French DECODA for the multiling challenge~\cite{favre-etal-2015-call}. 
The first baseline was based on Maximal Marginal Relevance (Baseline-MMR), the second baseline was the first words of the longest turn in the conversation, up to the length limit (Baseline-L), while the third baseline is the words of the longest turn in the first 25\% of the conversation, which usually corresponds to the description of the caller’s problem (Baseline-LB). Those baselines are described in more details in \cite{trione-2014-extraction}. 
Regarding previous computational models, the LIA-RAG system \cite{linharespontes:hal-01779304} was based on Jensen-Shannon (JS) divergence and TF-IDF approach. We did not find any information about the NTNU system.
We did not find more recent results or study on the DECODA dataset. 

Overall, in the challenge, the baselines were very difficult to beat with a ROUGE-2 of $4.0$. This quite low score shows the great difficulty of the task. Indeed, the transcriptions contain overlapping turns, fillers, pauses, noise, partial words, etc. Furthermore translators were required to translate speech phenomena such as disfluencies as closely as possible to the source language maintaining ‘naturalness’ in the target language. 

Comparing the performance of different pre-trained models fine-tuned on the DECODA summarization task, BARThez outperformed other models, with a ROUGE-2 score of $16.85$ using the \textit{Spk} data version. 

In general, all pre-trained models outperform previous results. BART-based pre-trained models (BARThez, mBART, mBARThez) have better performance than mT5 and BelGPT-2.

The difference between \textit{Spk} and \textit{NoSpk} is not significant. For some models like BARThez and mBART, \textit{Spk} (using speakers' prefix) performs slightly better than \textit{NoSpk}; while for mT5, \textit{NoSpk} (without the speakers' prefix) performs better than \textit{Spk}.

\subsection{Qualitative Analysis}

\begin{table*}[!h]
\begin{center}
\begin{tabularx}{\textwidth}{|l|X|}

      \hline
      \multicolumn{2}{|c|}{FR\_20091112\_RATP\_SCD\_0826}\\
      \hline
      References & Request for train schedules from Meaux station to Gare de l'Est at a given time. (1) \\
      & schedules RER E from Meaux to Gare de l'Est (166) \\
      \hline
      \multicolumn{2}{|c|}{\textbf{Spk}} \\
      \hline
      BARThez & Request for train schedules to go from Meaux station to Paris.  \\
      \hline
      BelGPT-2 & Request for information on the destination of the {\color{red}Meaux-Chappes-Saint-Lazare subway line}. {\color{red}Bus transfer at Saint-Lazare station. Is the call possible. Is the call possible to do it. Is the call possible. Is the call possible. Is the call}\\ 
      \hline
      \hline
      mBART & Request for SNCF train schedules to go to Paris from the station of Meaux {\color{red}from the station of Gare de Lyon in Paris}. Request for timetables to go {\color{red}to the station of Meaux at sixteen hours forty-nine} by leaving the station of Meaux at sixteen hours forty-nine {\color{red}by leaving the station of Gare de Lyon}. \\
      \hline
      mBARThez & train schedules in Île-de-France \\
      \hline
      mT5 & Request for timetable to {\color{red}go to the station of Meaux}. \\
      \hline
      \multicolumn{2}{|c|}{\textbf{NoSpk}} \\
      \hline
      BARThez & Request for train schedules to go to Paris. \\
      \hline
      BelGPT-2 &  , the journey from the station of Meaux to the {\color{red}Gare de Lyon, passing by the station of Bercy ) and finally the journey from the station of Bercy to the station of Bercy, passing by the station of Bercy, the journey from the station of Bercy to the station of Bercy, and finally the journey from the station of Bercy to the station of Montparnasse, the journey from the station of Bercy} \\
      \hline
      \hline
      mBART & Request for schedules for an RER line between Meaux {\color{red}Meaux} station and {\color{red}Gare de Lyon} station. {\color{red}Contact with the concerned service.} \\
      \hline
       mBARThez & RER {\color{red}B} schedules to go to Paris from Meaux station  \\
      \hline
      mT5 & Request for timetable to {\color{red}go to the station of Meaux}. \\
      \hline

\end{tabularx}
\caption{(Translation in English) Doc \textit{FR\_20091112\_RATP\_SCD\_0826} from DECODA’s test set, and associated reference and generated summaries by fine-tuning different models. Errors are marked in red.}
\label{table:english_example_decoda_summaries}
 \end{center}
\end{table*}

We observed that the BARThez pre-trained model fine-tuned on the \textit{Spk} version of DECODA led to the best ROUGE. We then randomly selected 20 test samples and performed an error analysis of the generated predictions, comparing them to references.

The references were written by different annotators and not all of them summarized the dialogue in the same way. Furthermore, because synopses are used to quickly convey information about the conversation between the caller and the RATP agent, they are even more abstract than typical dialogue summaries.

After comparing the synopses generated by BARThez (\textit{Spk}) to the references written by annotators, we observed several common types of errors:

\begin{itemize}
    \item \textbf{Hallucination:} This is a common type of error in neural NLG and is widely present in our generated synopses, meaning that the generated synopses are inconsistent or unrelated to the system input. 
    \item \textbf{Omission:} The other most common type of error, which means that some of the information presented in the references is not presented in the generated synopses.
    \item \textbf{Grammatical Error:} In our case, the generated synopses are generally smooth, but they may have incorrect syntax or semantics.

\end{itemize}

We categorize the errors and count their frequencies in Table~\ref{table: error_types}.

\begin{table}[!h]
\begin{center}
\begin{tabularx}{\columnwidth}{|l|X|}

      \hline
      Type of errors & Occurrences \\
      \hline
      Hallucination & 15 / 20 \\
      \hline
      Omission &  15 / 20 \\
      \hline
      Grammatical error & 5 / 20 \\
      \hline

\end{tabularx}

\caption{The most common error types of the BARThez (\textit{Spk}) model compared to the golden reference over 20 sampled dialogues, with the number of occurrences for each error type.}
\label{table: error_types}
\end{center}
\end{table}

We noticed that 7 of the 20 generated synopses have a \textit{Communication of the relevant service number} at the end, which is not presented in the reference or in the input dialogue. Since we truncated the length of the input sequences to a maximum of 1024 tokens, this could be the result of truncation: the lack of information in the final part of the dialogue causes system to sometimes generate a general sentence.

In addition, for each synopsis generated, we counted the number of errors. Note that for the same type of errors, we can count more than once. For example, for omission, if two different points of information are missing from the generated synopsis, we count twice. The results of the number of errors are reported in Table~\ref{table: number_error_frequence}, the total number of errors counts from 1 to 5 for above 20 randomly selected synopses.

\begin{table}[!h]
\begin{center}
\begin{tabularx}{\columnwidth}{|l|X|}

      \hline
       Number of errors & Synopses (total: 20) \\
      \hline
      1 & 4  \\
      \hline
      2 & 3 \\
      \hline
      3 & 10 \\
      \hline
      4 & 2 \\
      \hline
      5 & 1 \\
      \hline

\end{tabularx}
\caption{Number of synopses generated by BARThez (\textit{Spk}) with a total number of errors from 1 to 5.}
\label{table: number_error_frequence}
 \end{center}
\end{table}

We noticed that half of the synopses generated contain 3 errors, 1 sample contains 5 errors and 4 synopses contain only 1 error. For example, compared with the reference \textit{demande de précisions sur la tarification du métro, mais l'appelant, un particulier, est passé directement par le numéro VGC, refus du conseiller de le prendre en charge comme grand compte, indication du numéro 3246 pour qu'il rappelle}, the generated synopsis \textit{Demande de renseignement sur les tarifs RATP sur Paris ou la zone parisienne. Communication du numéro de téléphone.} has 3 errors : 
\begin{itemize}
    \item 1) omission - \textit{mais l'appelant, un particulier, est passé directement par le numéro VGC}; 
    \item 2) omission - \textit{refus du conseiller de le prendre en charge comme grand compte};
    \item 3) omission - \textit{indication du numéro 3246 pour qu'il rappelle}
\end{itemize}

Besides, while doing error analysis, we also noticed different writing styles of different annotators. For example, some golden references always begin with \textit{ask for (Demande de)} while some other references are much shorter, for example \textit{lost phone $>$ but not found (perdu téléphone $>$ mais pas retrouvé)}.

Taking \textit{FR\_20091112\_RATP\_SCD\_0826} from DECODA’s test set as an example, we compared the summaries generated by fine-tuning different pre-trained models with the references in Table~\ref{table:english_example_decoda_summaries} (translated in English, original French version and the dialogue in Appendix). Errors are marked in red.

For BARThez (\textit{Spk}) which got the highest ROUGE-2 score, the synopsis generated for this test example is quite good, however BARThez (\textit{NoSpk}) does not mention \textit{from Meaux station}. 

In general, for fine-tuned BelGPT-2, the generated synopses are correct in some respects but suffer from factual inconsistency. The model started to generate repetitive word sequences quickly. Its poor performance may be due to the type of model (decoder only) and its small size. 

mBART (\textit{Spk}) was good in the beginning, but it started to produce incorrect and unrelated information later on. As for mBART (\textit{NoSpk}), it incorrectly predicted \textit{Gare de Lyon station}
instead of \textit{Gare de l'EST}.

As for fine-tuned mBARThez (\textit{Spk}), it seems to be correct but missed some information like \textit{from Meaux to Gare de l'Est}. mBARThez (\textit{NoSpk}) wrongly predicted \textit{RER B} instead of \textit{RER E}.

mT5 wrongly predicted the direction, it should be \textit{from Meaux station to Gare de l'Est} but it wrongly took \textit{to go to the station of Meaux}. In addition, the generated synopses lack other information such as \textit{request for train schedules}, \textit{at a given time} and \textit{to Gare de l'Est}.

\section{Discussion and Conclusion}

Our experimental results show that the BARThez models provide the best performance, far beyond the previous state of the art on DECODA. However, even though the pre-trained models perform better than traditional extractive or abstractive methods, their outputs are difficult to control and suffer from hallucination and omission, there are also some grammatical errors.

On the one hand, the limit of maximum token length for pre-trained models evokes challenges to summarize such non-trivial dialogue. For long dialogue summarization, we can use retrieve-then-summarize pipeline or use end-to-end summarization models \cite{zhang-etal-2021-exploratory-study}. \cite{zhu-etal-2020-hierarchical} designed a hierarchical structure named HMNet to accommodate long meeting transcripts and a role vector to depict the difference between speakers. \cite{Longformer2020} introduced the Longformer with an attention mechanism that scales linearly with sequence length, making it easy to process documents of thousands of tokens or longer. However, these end-to-end summarization models capable of dealing with long sequences problems are both pre-trained in English language. 

On the other hand, pre-trained models were mostly learned on news corpora but far from the spontaneous conversation (repetitions, hesitation, etc.). Dialogue summarization therefore remains a challenging task.

Besides, in call center conversations scenarios, the caller (or user) might express their impatience, complaint or other emotions, we would like to further explore DEOCDA corpus for emotion-related dialogue summarization.

\section{Acknowledgements}

This research has been conducted within the THERADIA project supported by the Banque Publique d'Investissement (BPI) and the Association Nationale de la Recherche et de la Technologie (ANRT), under grant agreement No.2019/0729, and has been partially supported by MIAI@Grenoble-Alpes (ANR-19-P3IA-0003) and by COST Action Multi3Generation, CA18231 supported by COST (European Cooperation in Science and Technology).

\section*{Appendix}
\label{sec:appendix}

Table~\ref{table:example_decoda_summaries_fr} presents the references and summaries generated by fine-tuning different models of the test example \textit{FR\_20091112\_RATP\_SCD\_0826}, \textit{Spk} and \textit{NoSpk} versions of the dialogue are also shown below.

\textbf{\textit{Spk} version:} This version presents the conversations with the speakers' information.

      $<$Spk\_A$>$ bonjour \\
      $<$Spk\_B$>$ oui bonjour madame \\
      $<$Spk\_B$>$ je vous appelle pour avoir des horaires de train en+fait c' est pas pour le métro je sais pas si vous pouvez me les donner ou pas  \\
      $<$Spk\_A$>$ trains SNCF  \\
      $<$Spk\_B$>$ oui trains SNCF oui  \\
      $<$Spk\_A$>$ vous prenez quelle ligne monsieur  \\
      $<$Spk\_B$>$ euh la ligne euh enfin en+fait c' est pas SNCF enfin c' est Île-de-France quoi je sais pas comment ils appellent ça  \\
      $<$Spk\_B$>$ RER voilà c' est pour les RER  \\
      $<$Spk\_B$>$ voilà et euh je prends la ligne euh Meaux de Meaux pour aller à Paris je sais plus c' est laquelle c' est la  \\
      $<$Spk\_B$>$ E je crois  \\
      $<$Spk\_A$>$ d'accord et vous voulez donc les horaires euh pour quel jour et à quel moment monsieur \\ $<$Spk\_B$>$ euh là pour euh tout à l' heure euh pour euh aux environs de dix-sept heures en partant de la gare de Meaux \\
      $<$Spk\_B$>$ euh vers la Gare+de+l'Est à Paris \\
      $<$Spk\_A$>$ alors vous partez de Meaux et vous allez donc à la Gare+de+l'Est $<$Spk\_A$>$ et vous voudriez les horaires $<$Spk\_B$>$ voilà \\
      $<$Spk\_A$>$ vers euh dix-sept heures $<$Spk\_A$>$ alors moi je peux regarder ce+que $<$Spk\_B$>$ ouais dix-sept heures ouais \\
      $<$Spk\_A$>$ j' ai comme horaires un instant monsieur $<$Spk\_A$>$ s'il+vous+plaît $<$Spk\_B$>$ d'accord il y a \\
      $<$Spk\_B$>$ pas de souci \\
      $<$Spk\_A$>$ monsieur s'il+vous+plaît \\
      $<$Spk\_B$>$ oui \\
      $<$Spk\_A$>$ donc voilà ce+que j' ai comme horaires moi vous avez donc un départ à la gare de Meaux donc à seize heures quarante-neuf \\
      $<$Spk\_B$>$ seize heures quarante-neuf d'accord \\
      $<$Spk\_A$>$ et après vous avez celui de dix-sept heures dix-neuf \\
      $<$Spk\_B$>$ alors seize heures quarante-neuf dix-sept heures dix-neuf $<$Spk\_A$>$ OK d'accord $<$Spk\_B$>$ oui \\
      $<$Spk\_B$>$ ben je vous remercie \\
      $<$Spk\_A$>$ mais je vous en prie $<$Spk\_B$>$ bonne journée madame \\
      $<$Spk\_A$>$ merci à vous aussi monsieur au+revoir \\
      $<$Spk\_B$>$ au+revoir

\textbf{\textit{NoSpk} version:} This version presents the conversations without the speakers' information.

bonjour. oui bonjour madame. je vous appelle pour avoir des horaires de train en+fait c' est pas pour le métro je sais pas si vous pouvez me les donner ou pas. trains SNCF. oui trains SNCF oui. vous prenez quelle ligne monsieur. euh la ligne euh enfin en+fait c' est pas SNCF enfin c' est Île-de-France quoi je sais pas comment ils appellent ça. RER voilà c' est pour les RER. voilà et euh je prends la ligne euh Meaux de Meaux pour aller à Paris je sais plus c' est laquelle c' est la. E je crois. d'accord et vous voulez donc les horaires euh pour quel jour et à quel moment monsieur. euh là pour euh tout à l' heure euh pour euh aux environs de dix-sept heures en partant de la gare de Meaux. euh vers la Gare+de+l'Est à Paris. alors vous partez de Meaux et vous allez donc à la Gare+de+l'Est et vous voudriez les horaires voilà. vers euh dix-sept heures alors moi je peux regarder ce+que ouais dix-sept heures ouais. j' ai comme horaires un instant monsieur s'il+vous+plaît d'accord il y a. pas de souci. monsieur s'il+vous+plaît. oui. donc voilà ce+que j' ai comme horaires moi vous avez donc un départ à la gare de Meaux donc à seize heures quarante-neuf. seize heures quarante-neuf d'accord. et après vous avez celui de dix-sept heures dix-neuf. alors seize heures quarante-neuf dix-sept heures dix-neuf OK d'accord oui. ben je vous remercie. mais je vous en prie bonne journée madame. merci à vous aussi monsieur au+revoir. au+revoir.

\begin{table*}[!h]
\begin{center}
\begin{tabularx}{\textwidth}{|l|X|}

      \hline
      \multicolumn{2}{|c|}{FR\_20091112\_RATP\_SCD\_0826}\\
      \hline
      References & Demande d’horaires de train de la gare de Meaux à la gare de l'Est à une heure donnée. (1) \\
      & horaires RER E de Meaux à la Gare de l'Est (166) \\
      \hline
      \multicolumn{2}{|c|}{\textbf{Spk}} \\
      \hline
  
      BARThez & Demande d'horaires de train pour aller de la gare de Meaux à Paris. \\
      \hline
    
      BelGPT-2 & demande de renseignement sur la destination de la ligne de {\color{red}métro de Meaux-Chappes-Saint-Lazare}. {\color{red}Transfert de bus à la gare de Saint-Lazare. L'appel est-il possible. L'appel est-il possible de le faire. L'appel est-il possible. L'appel est-il possible. L'appel est} \\ 
      \hline
      \hline
      
      mBART & Demande d'horaires de trains SNCF pour aller à Paris en partant de la gare de Meaux {\color{red}en partant de la gare de Gare de Lyon à Paris}. Demande des horaires pour se rendre {\color{red}à la gare de Meaux à seize heures quarante-neuf} en partant de la gare de Meaux à seize heures quarante-neuf {\color{red}en partant de la gare de Gare de Lyon}. \\
      \hline
      mBARThez & horaires de train en Île-de-France \\
      \hline
      mT5 & Demande de horaires pour aller {\color{red}à la gare de Meaux}. \\
      \hline
      \multicolumn{2}{|c|}{\textbf{NoSpk}} \\
      \hline
      
      BARThez & Demande d'horaires de train pour aller à Paris. \\
      \hline
    
      BelGPT-2 &  , le trajet de la gare de Meaux vers la {\color{red}Gare de Lyon, en passant par la gare de Bercy ) et enfin le trajet de la gare de Bercy à la gare de Bercy, en passant par la gare de Bercy, le trajet de la gare de Bercy vers la gare de Bercy, et enfin le trajet de la gare de Bercy vers la gare de Montparnasse, le trajet de la gare de Bercy} \\
      \hline
      \hline
      
      mBART & Demande d'horaires pour une ligne de RER entre la gare de Meaux {\color{red}de Meaux} et la gare de {\color{red}Gare de Lyon}. {\color{red}Mise en relation avec le service concerné.} \\
      \hline
      mBARThez &  horaires du RER {\color{red}B} pour aller à Paris en partant de la gare de Meaux \\
      \hline
      mT5 & Demande de horaires pour {\color{red}aller à la gare de Meaux}. \\
      \hline

\end{tabularx}
\caption{Doc \textit{FR\_20091112\_RATP\_SCD\_0826} from DECODA’s test set, and associated reference and generated summaries by fine-tuning different models. Errors are marked in red. }
\label{table:example_decoda_summaries_fr}
 \end{center}
\end{table*}

\bibliographystyle{model5-names}
\bibliography{lrec2022-example}

\end{document}